\documentclass[12pt, a4paper]{article}
\usepackage{jheppub}

\usepackage{amsmath,amssymb}
\usepackage{xcolor}
\usepackage{comment}
\usepackage{setspace}

\newcommand{\cZ}{{\cal Z}}

\newcommand{\tp}{{\tilde p}}
\newcommand{\tv}{{\tilde v}}

\newcommand{\be}{\begin{eqnarray}}
\newcommand{\ee}{\end{eqnarray}}

\newcommand{\nt}{\notag\\}

\title{Feature extraction of machine learning and phase transition point of Ising model}

\author{Shotaro Shiba Funai}

\affiliation{Physics and Biology Unit, Okinawa Institute of Science and Technology (OIST),\\1919-1 Tancha Onna-son, Kunigami-gun, Okinawa 904-0495, Japan}

\emailAdd{shotaro.funai@oist.jp}

\abstract{
We study the features extracted by the Restricted Boltzmann Machine (RBM)
when it is trained with spin configurations of Ising model at various temperatures.
Using the trained RBM, we obtain the flow of iterative reconstructions (RBM flow) of the spin configurations
and find that in some cases 
the flow approaches the phase transition point $T=T_c$ in Ising model.
Since the extracted features are emphasized in the reconstructed configurations, 
the configurations at such a fixed point should describe 
nothing but the extracted features.
Then we investigate the dependence of the fixed point on various parameters and conjecture the condition where the fixed point of the RBM flow is at the phase transition point.
We also provide supporting evidence for the conjecture by analyzing the weight matrix of the trained RBM.
}

\begin{document}

\maketitle

\section{Introduction}

These days machine learning is studied and applied 
in various fields of research, 
and its technology is rapidly developed.
For example, image and video recognition is improved 
by the convolutional neural network~\cite{CNN,VGG},
and linguistic recognition progresses using 
deep neural network with the transformer~\cite{transformer}.
In any cases, one of the important goals is 
to train a machine so that it outputs 
good vector representations which describe
various essential features of input data.
However, how the machine extracts such features is 
not fully understood in a theoretical way.

In such theoretical research,
spin configurations in the Ising model are often used 
as image data for machine learning.
This enable us to 
discuss the extracted features using concepts of physics.
Especially, two-dimensional Ising model is the simplest statistical model to exhibit the second order phase transition
at the critical temperature $T=T_c$~\cite{Ising}.
Then in many previous studies of the Ising model using machine learning~\cite{Ising-ML,Ising-ML2,Ising-ML3,Ising-ML4},
researchers discussed the relation of the extracted features and the phase transition.

Moreover, in order to understand the critical phenomena of the phase transition,
the renormalization is the most important concept in statistical physics~\cite{RG0}. 
For example, the phase transition point corresponds to an unstable fixed point of the renormalization group (RG) flow in the two-dimensional Ising model.
On the other hand, the feature extraction by machine learning 
is a kind of information compression,
which reminds us of the coarse-graining and the renormalization.
Therefore, 
many researchers have discussed whether the feature extraction is related to the renormalization in the Ising model~\cite{RG3,RG,RG1,RG2,RG4}.

The author also studied the relation of the feature extraction and the renormalization group (RG) flow of the Ising model
in the previous papers~\cite{Iso:2018yqu,ShibaFunai:2018aaw}.
We used the Restricted Boltzmann Machine (RBM)~\cite{RBM1,RBM2,RBM},
since it is one of the most suitable machine learning methods
for a dataset with the probability distributions.
Our dataset consists of spin configurations at various temperatures,
including both higher and lower than the critical temperature. 
After training the RBM with this dataset, 
we iteratively reconstruct the spin configurations using the trained RBM,
and obtain the flow of the probability distribution of configurations.
We named it the RBM flow,
and found that in some cases 
the fixed point of the RBM flow (which is called RBM fixed point)
appears around the critical temperature.
This is an interesting phenomenon 
since the RBM doesn't have any prior knowledge about the phase transition.
%
In such cases,
the RBM flow goes away from $T=0,\infty$
and approaches $T=T_c$.
This is exactly the opposite direction to the RG flow,
then it should be related to the inverse renormalization
and the super-resolution~\cite{SR,SR2,SR3}.

The RBM fixed point should represent nothing but 
the feature extracted by the RBM,
since the extracted features must be emphasized in the reconstructed configurations.
Then in Ref.\,\cite{Iso:2018yqu},
we suggested that the extracted feature in our cases
may be the scale invariance,
which is a notable property of configurations at the phase transition point.
However, the authors of Ref.\,\cite{CA} pointed out that,
while the RBM captured the existence of two phases,
the geometrical information in the configurations 
was learned by another neural network (NN) to measure
temperature of the configurations in the RBM flow.
This NN is called NN thermometer, 
which is trained by supervised machine learning 
so that it outputs correct temperature of input configurations.
They suggest that the scale invariance of the configurations
is a feature extracted by the NN thermometer, not by the RBM.

In this paper, to clarify this point,
we study the RBM flow without using the NN thermometer.
Instead, to measure temperature of the configurations,
we use the relation of temperature and energy
obtained by numerical calculations.
Then we discuss how the RBM fixed point depends on 
parameters in our dataset and the RBM.
Based on this analysis, we conjecture that 
\begin{itemize}
\item 
If the dataset contains configurations at higher temperature
(with the size of configurations fixed),
the RBM fixed point goes to higher temperature.
\item
However, if the size of configurations is large enough
(with the range of temperature fixed),
the RBM fixed point is around the phase transition point.
\end{itemize}
We will give its precise expression in the following sections,
and provide supporting evidence for this conjecture
by analyzing 
the weight matrix of the RBM and the features extracted by the RBM.

The paper is organized as follows. 
In section \ref{sec:2}, we explain how we generate spin configurations, make our dataset, train the RBM, and obtain the RBM flow.
In section \ref{sec:3}, we discuss the dependence of the RBM fixed point on important parameters, and conjecture the conditions where the RBM fixed point is at the phase transition point in the Ising model.
Finally, we conclude our discussion in section \ref{sec:4}.

\section{Ising configurations and RBM}
\label{sec:2}

In this paper, we concentrate on the two-dimensional ferromagnetic Ising model with no external field
and with interactions among only the nearest neighbor spins. 
The Hamiltonian is given as 
\begin{eqnarray} \label{eq:H}
{\cal H} = -\sum_{\langle i,j\rangle} \sigma_i \sigma_j
\end{eqnarray}
where 
$\sigma_i=\pm 1$ correspond to up/down spins
and 
the indices $i,j$ denote sites in the square lattice with periodic boundary condition.
$\langle i,j \rangle$ means 
the nearest neighbor pairs of sites.
Note that the interaction parameter is already fixed in Eq.\,(\ref{eq:H}) and 
we set the Boltzmann constant $k_B$ to be equal to 1 in the following,
therefore
all the physical quantities in this model are written as functions of only temperature $T$.

\subsection{Generating spin configurations}

\begin{figure}
    \centering
    \includegraphics[width=0.9\linewidth]{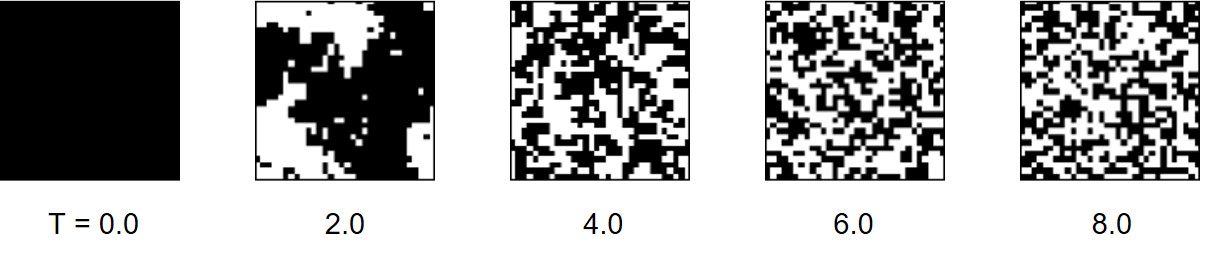}
    \caption{The spin configurations at temperature $T=0, 2, 4, 6, 8$ with size $N_v=32^2$
    generated by Metropolis Monte Carlo simulations.
    Configurations become random-like at higher temperature.}
    \label{fig:configs}
\end{figure}

We first construct samples of configurations of Ising model (\ref{eq:H}) as in Fig.\,\ref{fig:configs}.
In this paper, we use the configurations with the number of sites in the square lattice
\be
N_v = 7^2,\, 10^2,\, 20^2,\, 32^2.
\ee

The spin configurations at temperature $T$ are generated with the method of Metropolis Monte Carlo (MMC) simulation.
In this method, we first generate a random
configuration $\{\sigma_i\}$. 
Then we choose one of the spins $\sigma_i$ and flip its spin with the probability
\be
p = \min \left[ 1,\, e^{-dE_i/T} \right]
\ee
where $dE_i$ is the change of energy of this configuration by flipping.
After many iterations of flipping all the spins, the configuration
approaches the equilibrium distribution at $T$.
In our simulation, we flip the spins in $100N_v$ times to construct spin configurations,
and 
generate our dataset which includes the same number of configurations at $N_{temp}$ temperatures\footnote{
For $T=0$, we practically set $T=10^{-6}$ for numerical calculations.}
\be
T=0,\, 0.1,\, 0.2,\, \ldots,\, 0.1\times (N_{temp}-1),
\ee
where the number of configurations at each temperature is
\be\label{eq:Nconf}
N_{conf} = \min\left[ 2\times 10^3,\, 2\times \frac{10^5}{N_{temp}} \right].
\ee
Although such a dataset of spin configurations 
may be unnatural in physical systems, 
we choose them so that our dataset includes various image patterns from uniform ones to random ones.

\begin{figure}
    \centering
    \includegraphics[height=0.3\linewidth]{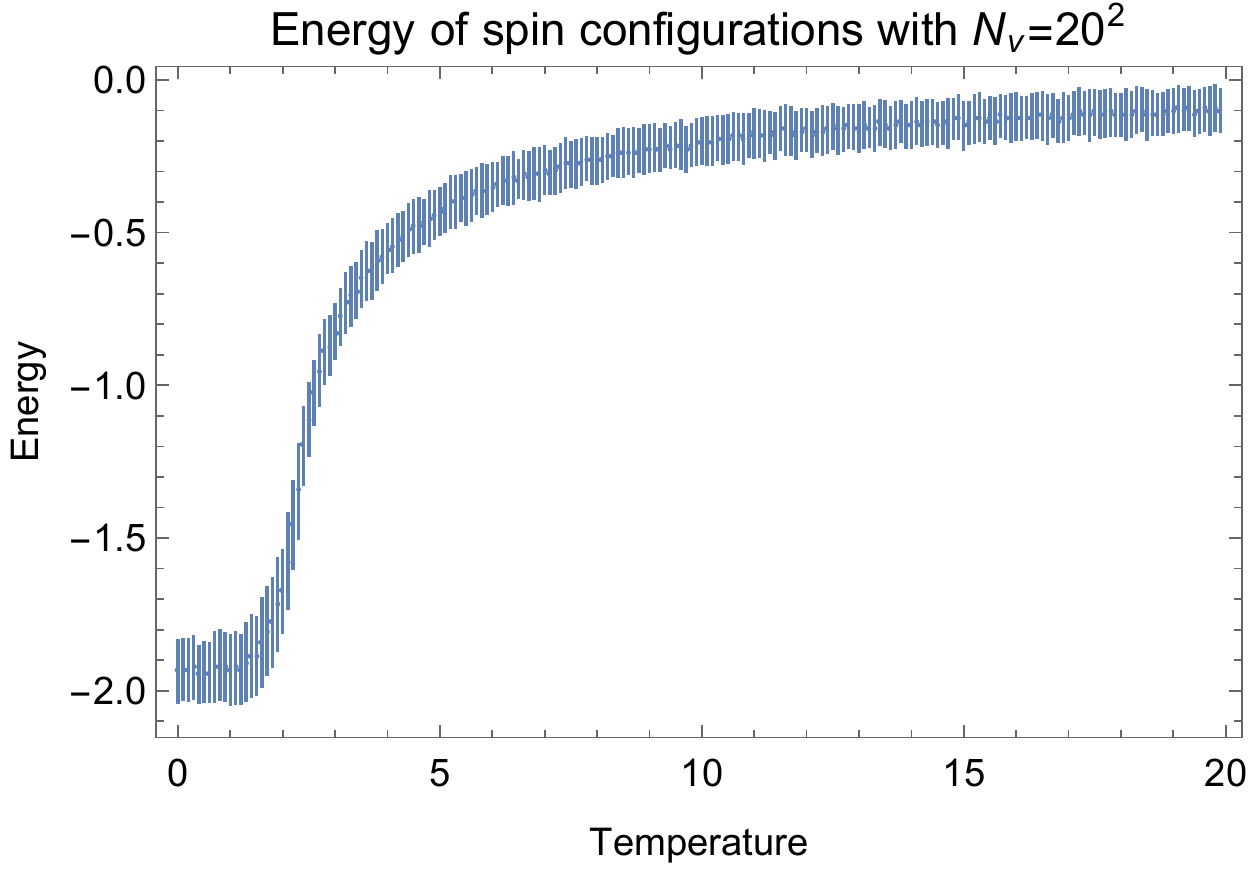}\quad
    \includegraphics[height=0.3\linewidth]{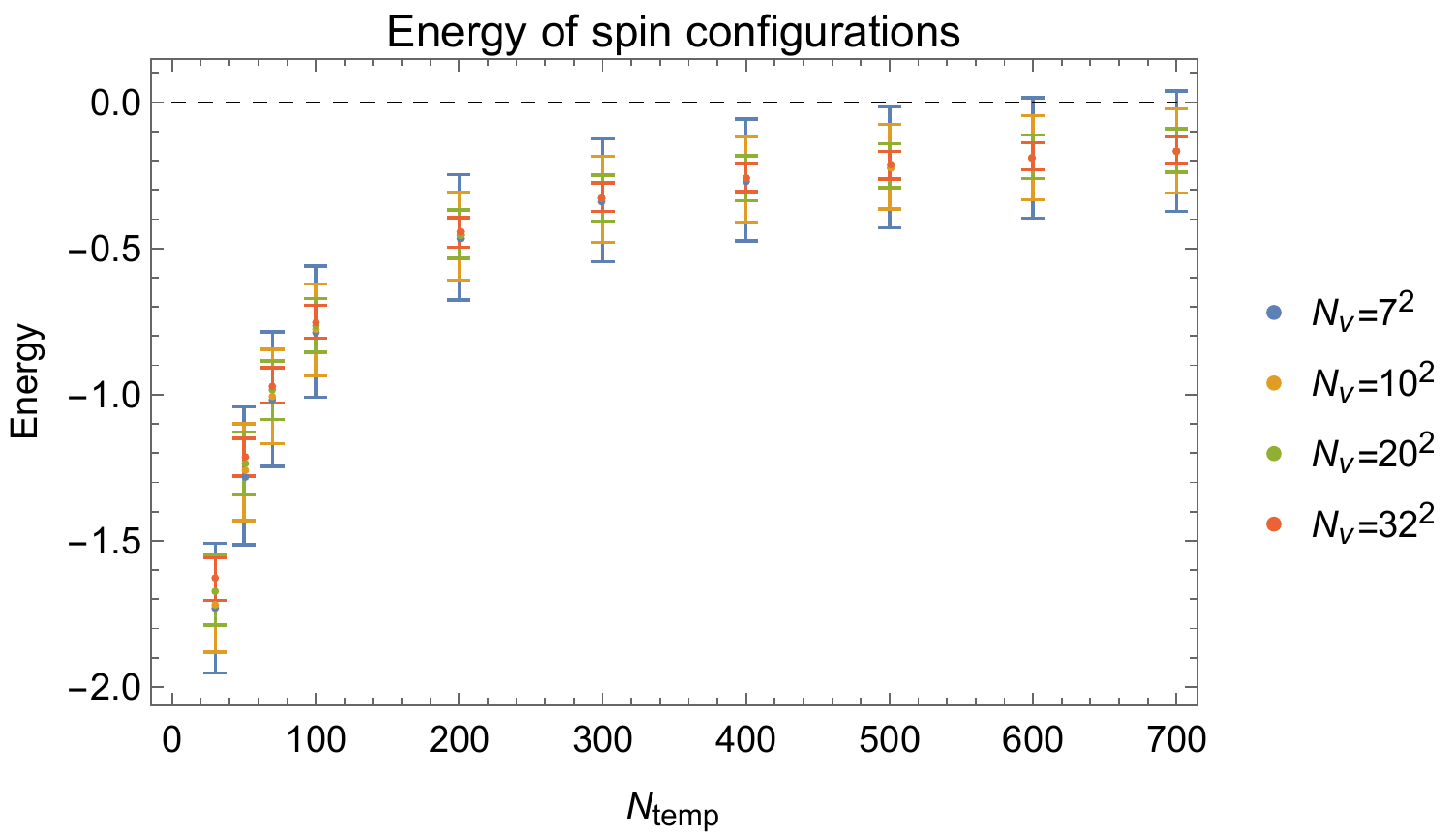}
    \caption{Energy per site of generated spin configurations 
    at each temperature $T$ with size $N_v=20^2$ (left) and 
    that of our dataset in each case of $N_v$ and $N_{temp}$ (right).
    All the error bars in this paper show the standard deviations.}
    \label{fig:en_configs}
\end{figure}

In the following analysis, we study the cases of 
\be
N_{temp}=30,\, 50,\, 70,\, 100,\, 200,\, 300,\, 400,\, 500,\, 600,\, 700.
\ee
Note that the maximum temperature $T_{max} = 0.1\times (N_{temp}-1)$ in all the cases
is higher than the critical temperature 
$T_c=2.27$,
which means our dataset always includes both configurations above $T_c$ and below $T_c$.
Figure \ref{fig:en_configs} shows the energy per site of the generated configurations 
at each temperature $T$ with size $N_v=20^2$ (in the left panel) and
that of our dataset in each case of $N_v$ and $N_{temp}$ (in the right panel).

\subsection{Training RBM}
\label{sec:2.2}

Once we fix the size of configurations $N_v$ and the number of temperatures $N_{temp}$,
we specify one of our datasets 
and obtain the probability distribution of configurations $q(\{\sigma_i\})$.
Then we choose the Restricted Boltzmann Machine (RBM) as one of the most suitable methods to learn the probability distributions of input data.

\begin{figure}[t]
    \centering
    \includegraphics[width=0.3\linewidth]{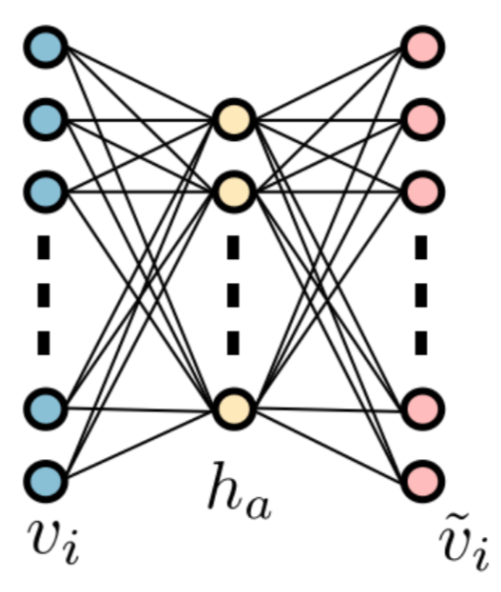}
    \caption{The neural network of RBM with a visible layer $\{v_i\}$ and a hidden layer $\{h_a\}$. 
    These two layers are coupled without intra-layer couplings. 
    The RBM generates the reconstructed configurations $\{\tv_i\}$
    from the input configurations $\{v_i\}$ through the hidden configurations $\{h_a\}$.}
    \label{fig:RBM}
\end{figure}

The RBM consists of the visible layer $v_i$ and the hidden layer $h_a$,
as shown in Fig.\,\ref{fig:RBM}.
The generated configurations $\{\sigma_i\}$ are input into the visible layer $\{v_i=\pm 1\}$,
which means that the number of neurons in the visible layer is equal to the size of configurations $N_v$, 
{\em i.e.,} $i=1,\ldots,N_v$.
On the other hand, the hidden layer can have an arbitrary number of neurons $N_h$. 
In this paper, we consider the cases of 
\be
N_h=1^2, 2^2, 3^2, \ldots, N_v,
\ee
thus the spin variables in the hidden layer are given as $\{h_a=\pm 1\}$ with $a=1,\ldots,N_h$.

Ideally, the RBM 
is trained so that it outputs the configurations with the same probability distribution as input data. 
The probability distribution of input configurations is given as $q(\{v_i=\sigma_i\})$,
while the probability distribution of output configurations is defined, 
using the ``energy'' function
\be
\Phi(\{v_i\},\{h_a\}) = -\sum_{i,a} v_i W_{ia} h_a - \sum_i b_i^{(v)} v_i - \sum_a b_a^{(h)} h_a\,,
\ee
by Boltzmann distribution in the process of 
$\{v_i\} \to \{h_a\}$ and $\{h_a\} \to \{\tv_i\}$:
\be
p(\{h_a\}) = \sum_{\{v_i\}} \frac{e^{-\Phi(\{v_i\},\{h_a\})}}{\cZ}\,,\quad
\tp(\{\tv_i\}) = \sum_{\{h_a\}} \frac{e^{-\Phi(\{\tv_i\},\{h_a\})}}{\cZ}\,.
\ee
Here $\{\tv_i\}$ is the final output of the RBM and we call them the reconstructed configurations.
$\cZ=\sum_{\{v_i,h_a\}}e^{-\Phi(\{v_i\},\{h_a\})}$ is the partition function.
The weight matrix $W_{ia}$ and the biases $b_i^{(v)}, b_a^{(h)}$ are 
parameters of the RBM which are optimized by training.

Practically, we train the RBM so as to 
minimize the distance between the probability distributions of 
input $q(\{v_i\})$ and output $\tp(\{\tv_i\})$ 
by optimizing the weight matrix and the biases.
The distance is defined as Kullback-Leibler (KL) divergence, 
or relative entropy, and given by
\be
{\rm KL}(q||\tp) := \sum_{\{v_i\}} q(\{v_i\}) \log \frac{q(\{v_i\})}{\tp(\{v_i\})}\,.
\ee
In other words, this is the loss function for training RBM.
The weights and the biases are optimized so that the loss function approaches
its local minimum.


To find the local minimum,
we use the method of stochastic gradient descent (SGD).
However, it is too difficult to evaluate the partition function $\cZ$
in realistic time,
then 
we avoid this difficulty by using the method of Gibbs sampling to approximately evaluate the gradients.
Practically we employ a more simplified method,
which is called the method of contrastive divergence (CD)~\cite{CD}:
we simply stop the process of iterative Gibbs samplings
at the fixed number of steps even before its convergence.
Especially, we adopt the simplest version of CD, called CD1.

In our training, 
for all the datasets with $N_v$, $N_h$, $N_{temp}$,
we set the learning rate $\epsilon=10^{-3}$ in SGD
and use the method of momentum with the parameter $\mu=0.5$
for rapid convergence.
We divide the dataset of configurations into training data and test data, and then train the RBM in $10^5$ epochs.
We check that the loss function of training data converges
at its local minimum and also that of test data doesn't increase,
which shows that the RBM is not overtrained by the training data.
Both training and test data have the same number of configurations, that is, $N_{conf}/2 = \min [10^3, 10^5/N_{temp}]$ configurations at each temperature.


After the training finished,
by using the optimized values of weights and biases,
we can calculate the expectation values of neurons as 
\be
\langle h_a \rangle = \tanh\left(\sum_i v_i W_{ia} + b^{(h)}_a\right) \nt
\langle \tv_i \rangle = \tanh\left(\sum_a W_{ia} h_a + b^{(v)}_i\right)\,.
\ee
Note that the reconstructed configurations $\{\tilde{v}_i=\pm 1\}$ are obtained
by replacing an expectation value $\langle \tv_i\rangle$
with a probability $(1\pm \langle \tv_i\rangle)/2$
so that the expectation value is kept unchanged.
Therefore, if $\langle \tv_i\rangle$ is not closed to $\pm 1$,
a kind of random selection occurs
and it causes random noise in the reconstructed configurations.

\subsection{RBM flow}

After the training finished, the probability distribution of input configurations $q(\{v_i\})$ and that of output (reconstructed) configurations
$\tp(\{\tv_i\})$ are similar but slightly different.
It is because the KL divergence is practically not zero
even after the training.
Then, if we input again the reconstructed configurations into the same RBM,
we obtain another probability distribution $\tilde \tp(\{\tv_i\})$ of the reconstructed configurations. 
Doing this reconstruction process iteratively, 
we obtain the flow of probability distribution of the spin configurations:
\be
q(\{v_i\})\to \tp(\{\tv_i\})\to \tilde\tp(\{\tv_i\})\to \cdots\,,
\ee
which we call the RBM flow~\cite{Iso:2018yqu}.

As we discussed in our previous papers~\cite{Iso:2018yqu,ShibaFunai:2018aaw},
the RBM flow has its fixed points in the parameter space of temperature $T$,
although there are no fixed points in the space of spin configurations.
To estimate temperature of the reconstructed configurations, 
we can use the following two ways:
\begin{itemize}
    \item 
    We train another neural network to output the correct value of temperature $T$ 
    ({\em i.e.,} the parameter of the MMC simulation) of input configurations. 
    Then we obtain the probability distribution of $T$ and regard $T$ at the peak of the distribution as the estimated temperature of input configurations.
    \item
    We calculate energy $E$ of configurations using Hamiltonian (\ref{eq:H})
    and estimate $T$ using the numerical relation of $T$ and $E$ 
    in the left panel of Fig.\,\ref{fig:en_configs}.
    Then we regard the averaged $T$ as the estimated temperature of the configurations.
\end{itemize}
We can check that these two methods give us the consistent results with each other.
In the previous study we used the first method,
but in this paper we used the second. 

Since the RBM extracts features of input data in the training process,
we can expect that the extracted features are emphasized in the reconstructed configurations
along the RBM flow and that its fixed points represent nothing but the extracted features. 
Therefore, in the next section, we study the fixed points of RBM flow in detail.

\section{Fixed point of RBM flow}
\label{sec:3}

In our previous papers, 
we showed several interesting properties of the RBM flow and its fixed points, which we call RBM fixed points.

In particular, the RBM flow approaches the phase transition point $T=T_c=2.27$
of the 2d Ising model (\ref{eq:H}), while goes away from $T=0,\infty$~\cite{Iso:2018yqu}.\footnote{
Training data: $10^3$ configurations with $N_v=10^2$ at each $T=0, 0.25, \ldots, 6$ ($H=0$).
}
This is exactly the opposite direction to the renormalization group (RG) flow,
although some researchers suggested that the feature extraction in unsupervised machine learning like the RBM may be a kind of coarse-graining and correspond to the RG flow~\cite{RG}.

Moreover, if the training data includes spin configurations in 2d Ising model
with external magnetic field $H\neq 0$,
the RBM flow approaches the points with maximal heat capacity in $(T,H)$ space~\cite{ShibaFunai:2018aaw}.\footnote{
Training data: $10^3$ configurations with $N_v=10^2$ at each $(T,H)$,
where $T=0, 0.5, \ldots, 9.5$ and $H=0, 0.5, \ldots, 4.5$.
}
These fixed points include $(T,H)=(T_c,0)$, 
then near this point we find again the opposite direction
of the RBM flow to the RG flow.
In the region far from this point,
the behavior of the RBM flow is also different from the RG flow.

This is an interesting but mysterious result.
The reason is not clarified yet:
It may be related to the scale invariance,
since it is an important property of spin configurations at $T=T_c$,
and we will discuss it in Sec.\,\ref{sec:3.4}.
The condition is also not clear, then we study the parameter dependence of the RBM fixed point in the next subsections.

\subsection{Dependence on $N_h$}
\label{sec:Nh}

\begin{figure}[t]
    \centering
    \includegraphics[width=0.7\linewidth]{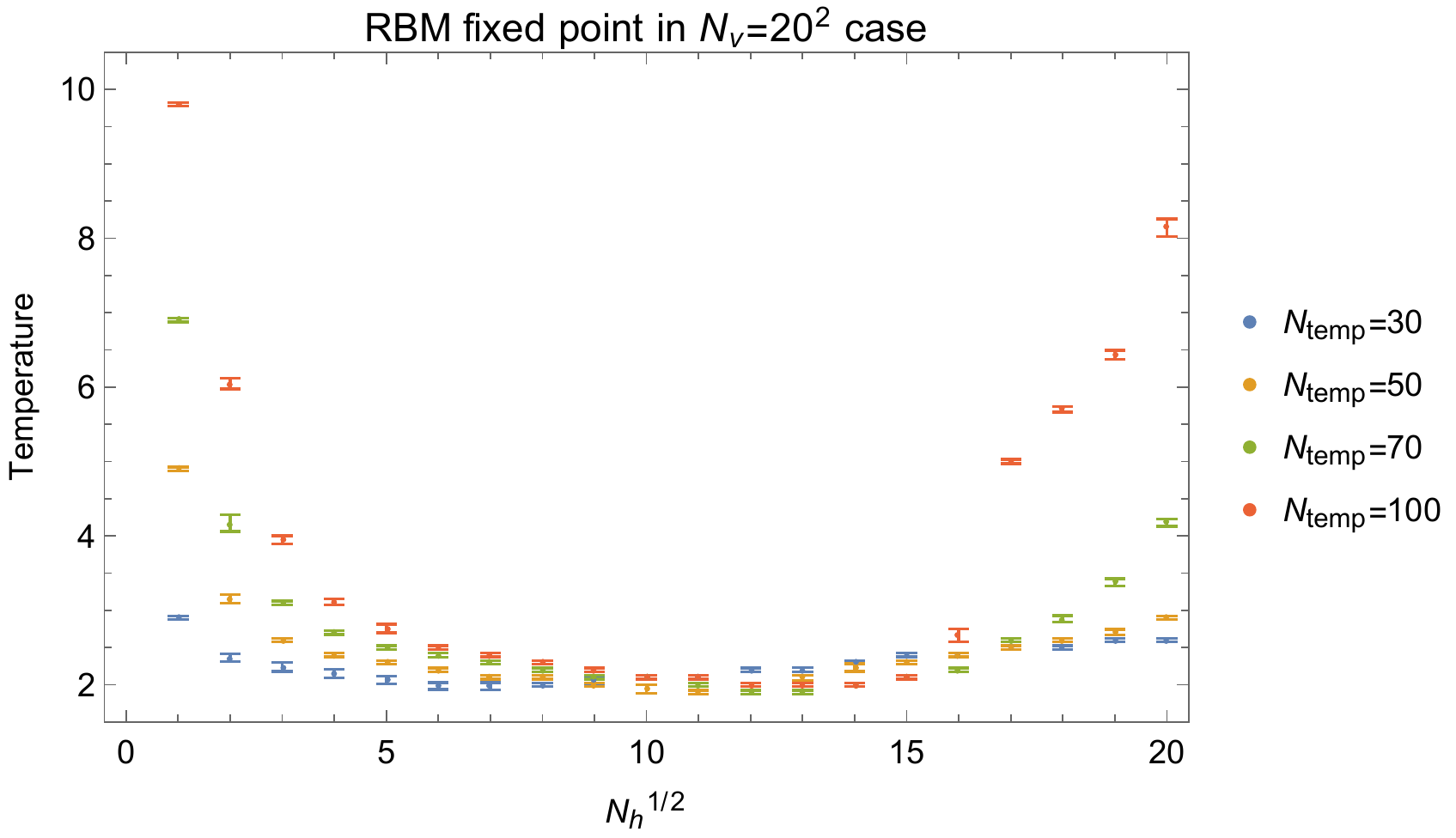}
    \caption{Temperature of RBM fixed point. 
    Around $\sqrt{N_h/N_v}\sim 1/2$, the fixed point is at the lowest temperature and around $T\sim T_c=2.27$.}
    \label{fig:RBM_temp}
\end{figure}

\begin{figure}[t]
    \centering
    \includegraphics[width=0.6\linewidth]{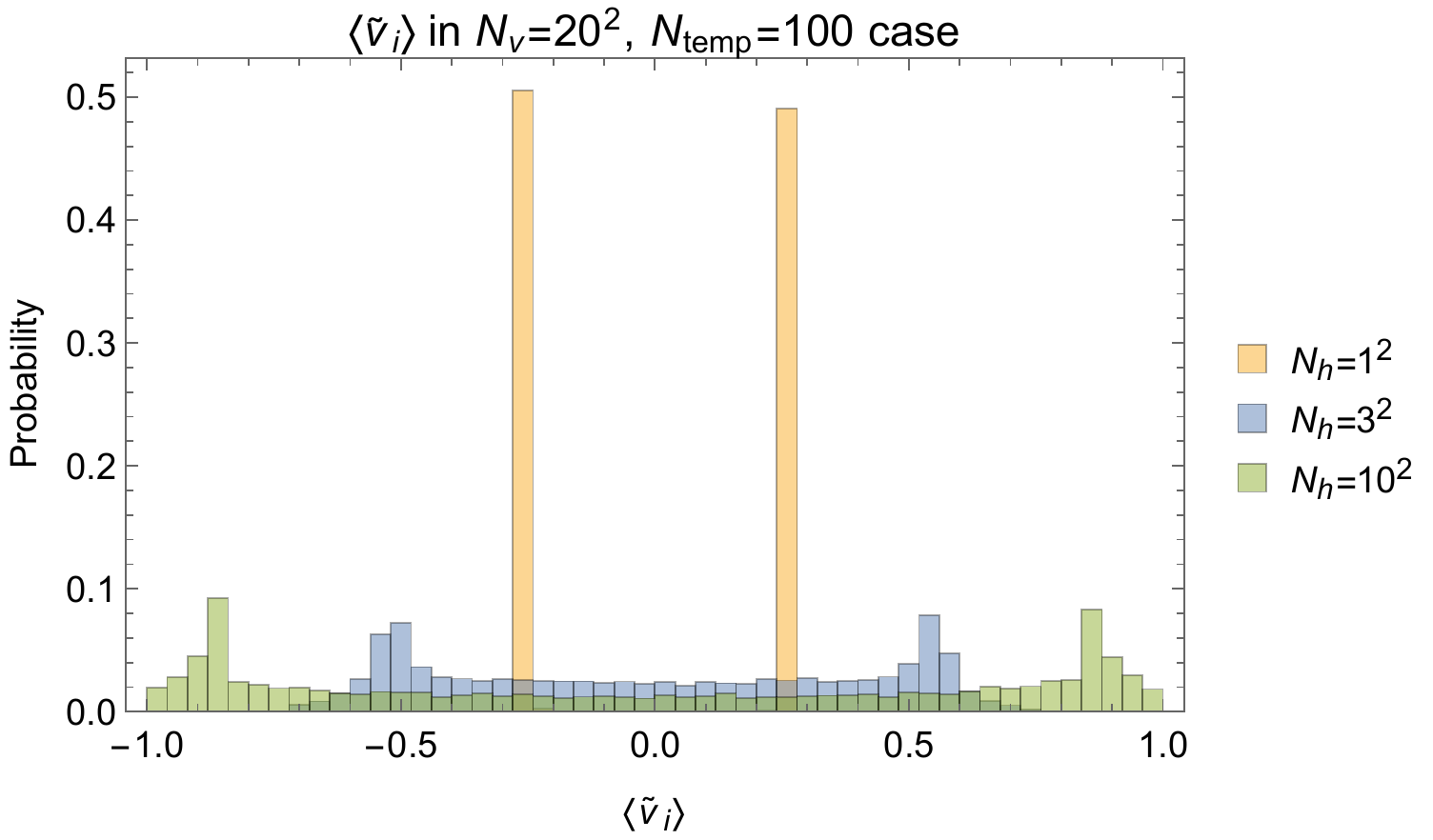}
    \caption{Probability distribution of the expectation value $\langle\tv_i\rangle$. For small $N_h$, $\langle\tv_i\rangle$ are not close to $\pm 1$, which causes random noise in the reconstructed configurations $\tv_i$.}
    \label{fig:prob_vi}
\end{figure}

First we study the dependence on the number of hidden neurons $N_h$.
The RBM fixed points in terms of temperature are shown in Fig.\,\ref{fig:RBM_temp}.

For small and large $N_h$, or $N_h/N_v\sim 0$ and $1$,
the RBM fixed point tends to be at high temperature.
We can check it especially for large $N_{temp}$.
This is understandable since
\begin{itemize}
    \item If $N_h/N_v\sim 0$, the RBM learns only few and unclear patterns, 
    and the expectation values $\langle \tv_i\rangle$ are not close to $\pm 1$,
    as shown in Fig.\,\ref{fig:prob_vi}.
    This causes random noise in the reconstructed configurations $\tv_i$, as mentioned in the end of Sec.\,\ref{sec:2.2}.
    \item If $N_h/N_v\sim 1$, the RBM learns many random-like patterns, 
    and such patterns appear in the reconstructed configurations $\tv_i$.
\end{itemize}

\begin{figure}[t]
    \centering
    \includegraphics[width=0.48\linewidth]{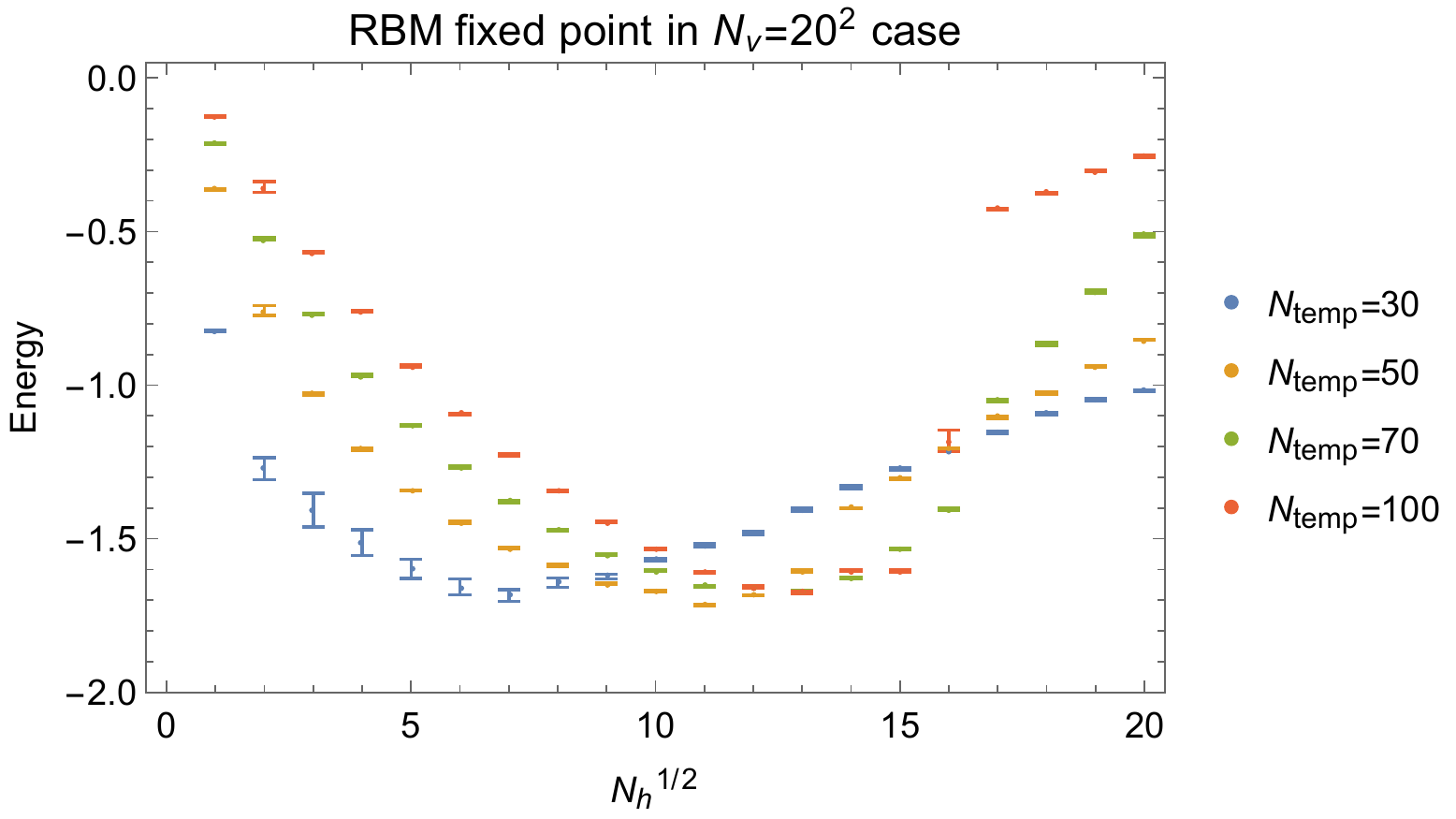}\quad
    \includegraphics[width=0.48\linewidth]{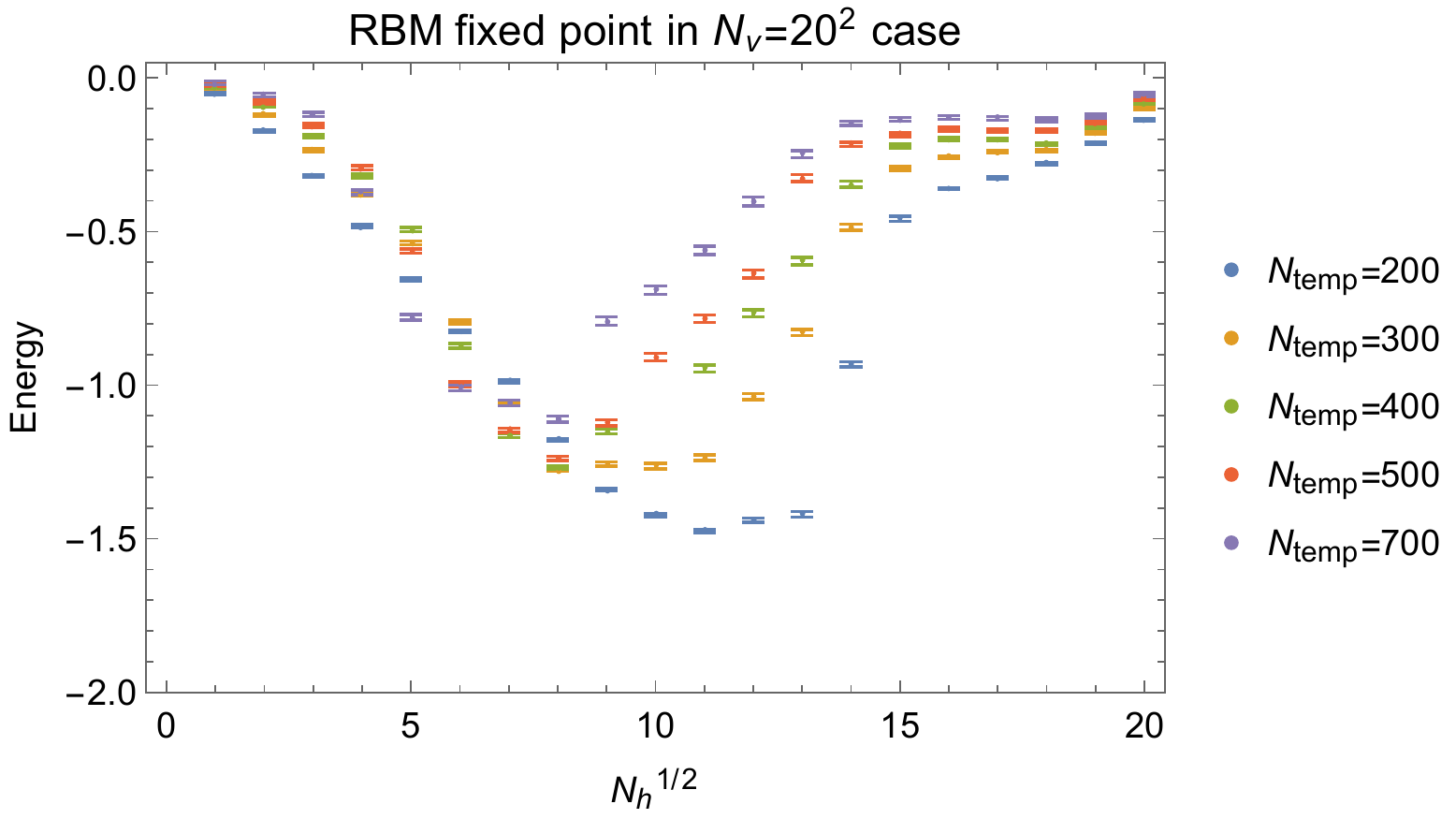}
    \caption{Energy of RBM fixed point. For $N_{temp}\leq 100$ (left panel), 
    $N_h$ with the minimum energy largely changes depending on $N_{temp}$.}
    \label{fig:RBM_en}
\end{figure}

Around $\sqrt{N_h/N_v}\sim 1/2$,
on the other hand, 
the RBM fixed point is at the lowest temperature 
and around $T\sim T_c=2.27$.
This is a consistent result with our previous papers~\cite{Iso:2018yqu,ShibaFunai:2018aaw}.
In order to study the RBM fixed points in this region, 
it is better to use energy instead of temperature
as in Fig.\,\ref{fig:RBM_en},
since the energy rapidly changes around $T\sim T_c$ 
(as shown in the left panel of Fig.\,\ref{fig:en_configs}).
Then we find that, 
especially for $N_{temp}\leq 100$ (in the left panel of Fig.\,\ref{fig:RBM_en}),
$N_h$ with the minimum energy of the RBM fixed point
largely changes depending on $N_{temp}$.

\begin{figure}
    \centering
    \includegraphics[width=0.45\linewidth]{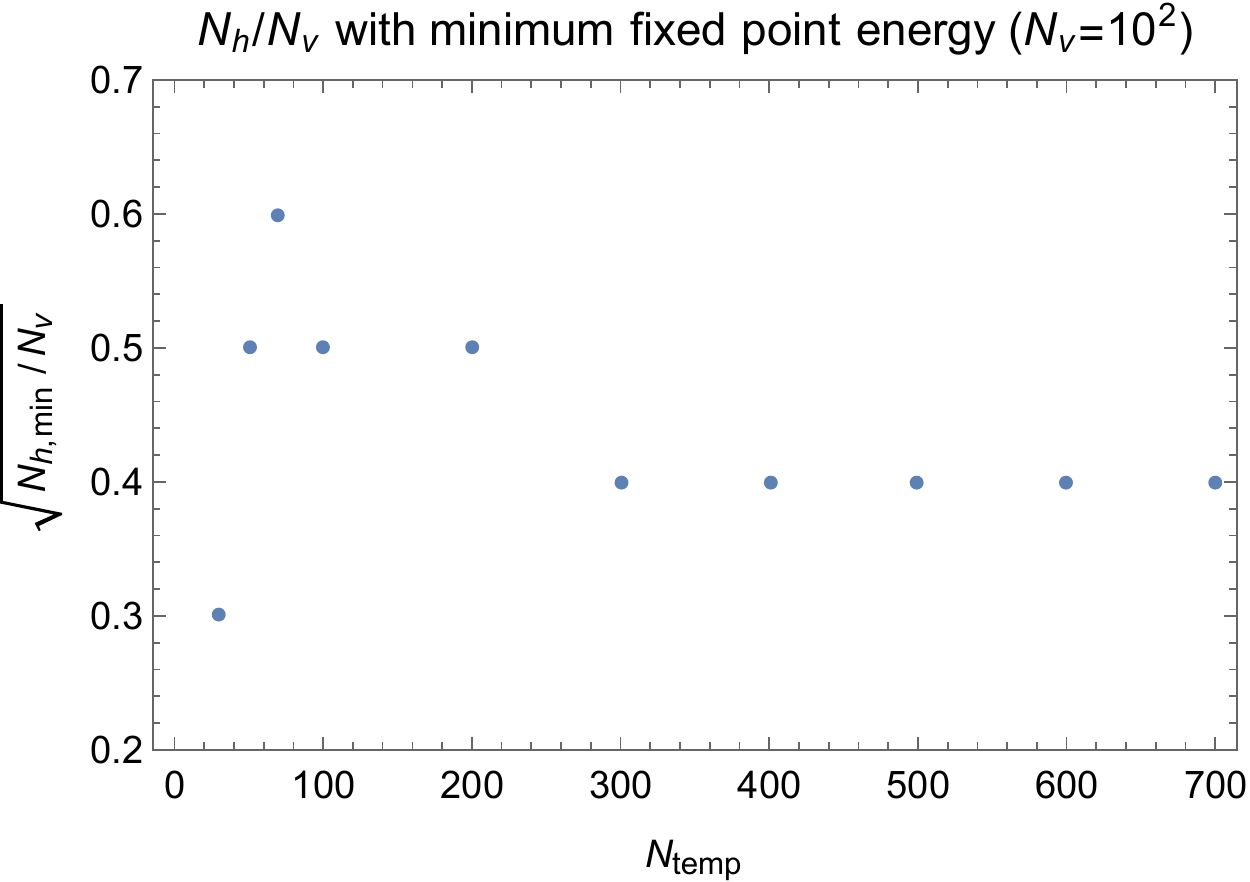}\quad
    \includegraphics[width=0.45\linewidth]{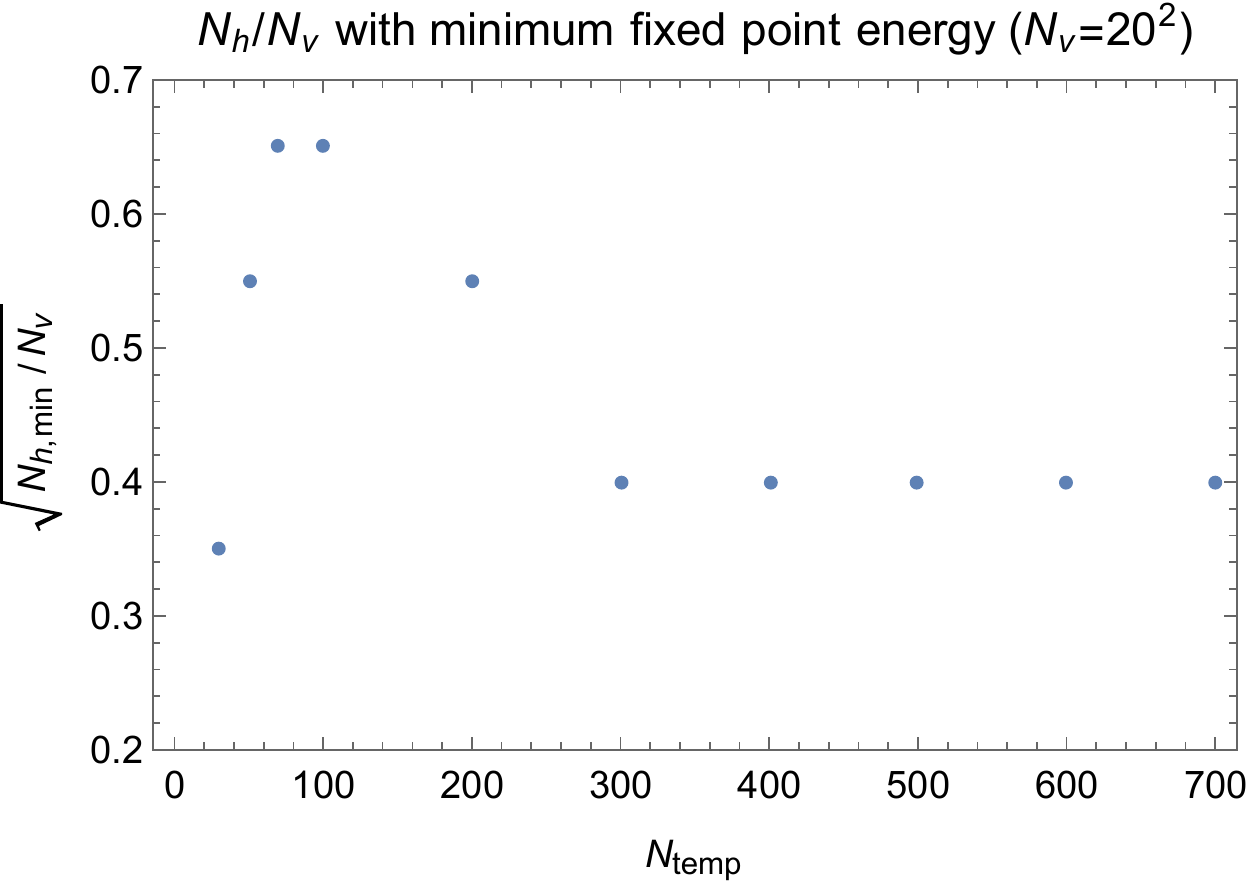}
    \caption{$N_h$ with minimum energy of RBM fixed point (which we call $N_{h,min}$).
    The value of $\sqrt{N_{h,min}/N_v}$ is constant for $N_{temp}\geq 300$ and doesn't depend on $N_v$.
    This enables us to discuss the case of large $N_{temp}, N_v$.}
    \label{fig:NhNv}
\end{figure}

Let us take a close look at 
$N_h$ with the ``minimum'' energy of the RBM fixed point,
which we call $N_{h,min}$ henceforth, 
although we have only discrete data points at $N_h=({\rm integer})^2$.
Figure~\ref{fig:NhNv} shows 
how $N_{h,min}$ depends on $N_{temp}$ and $N_v$.
Unlike for $N_{temp}\leq 100$, we find that
the value of $\sqrt{N_{h,min}/N_v}$ converges at $0.4$
for $N_{temp}\geq 300$,
and these values (0.4 and 300) don't depend on $N_v$.
Therefore, 
we can assume that 
$N_{h,min} = (0.4)^2 N_v$ for $N_{temp}\geq 300$.

Based on this observation, 
if we focus on 
the RBM fixed point at $N_h=N_{h,min}$
and study its dependence on $N_{temp}$ and $N_v$,
we can 
discuss the case of large $N_{temp}, N_v$ 
as in the next subsections.

\subsection{Dependence on $N_{temp}$ and $N_v$}

Next we concentrate on $N_h$ with the minimum energy of RBM fixed point, or $N_{h,min}$,
and study the dependence on the number of temperatures $N_{temp}$ and the size $N_v$.

\begin{figure}[t]
    \centering
    \includegraphics[width=0.7\linewidth]{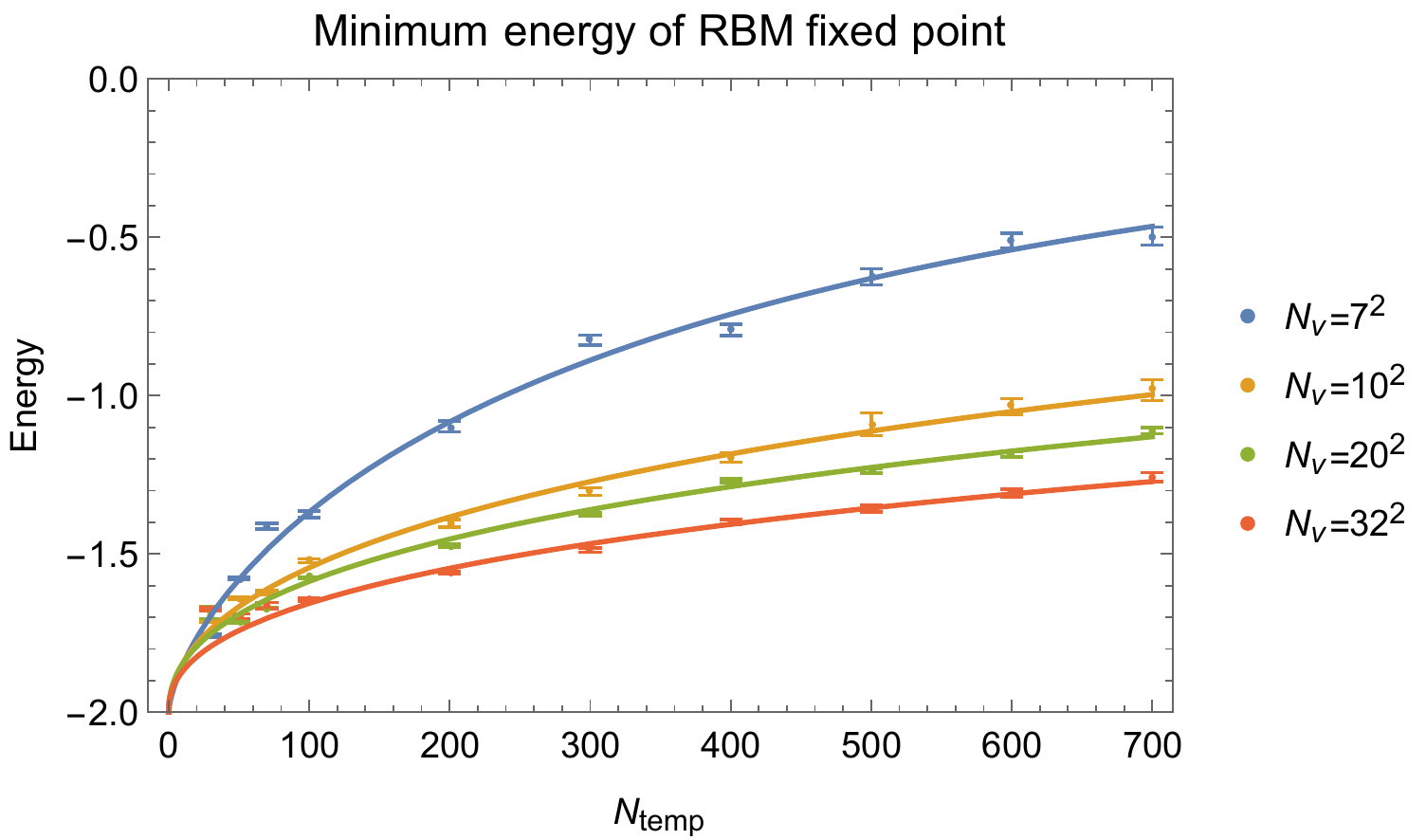}
    \caption{Minimum energy $E_{min}$ of RBM fixed point and its fitting function for each $N_v$.
    The fitting function is given in Eq.\,(\ref{eq:fit}).}
    \label{fig:min_en}
\end{figure}

Let us first discuss the $N_{temp}$ dependence.
Figure \ref{fig:min_en} shows that 
the minimum energy $E_{min}$ of the RBM fixed point (at $N_h=N_{h,min}$) for each $N_v$ 
is a monotonically increasing function of $N_{temp}$
and can be fitted to the function 
\be \label{eq:fit}
E_{min} = -2 \exp\left[ -aN_{temp}^b \right]
\ee
with the fitting parameters $a$ and $b$.
For this fitting,
we use only the data points with $N_{temp}\geq 100$.
The reason will be stated later in this subsection.
Here we note that 
$a\geq 0$ must be satisfied,
since the energy per site satisfies $-2\leq E\leq 0$ in the Ising model (\ref{eq:H}).
Also, $b\geq 0$ is satisfied
since $E_{min}$ is a monotonically increasing function of $N_{temp}$,
just like the averaged energy of training data
(in the right panel of Fig.\,\ref{fig:en_configs}).
Therefore, 
we can conjecture that the minimum energy approaches $E_{min} \to 0$ 
in the limit of $N_{temp}\to \infty$ with $N_v$ fixed.

\begin{figure}
    \centering
    \includegraphics[width=0.48\linewidth]{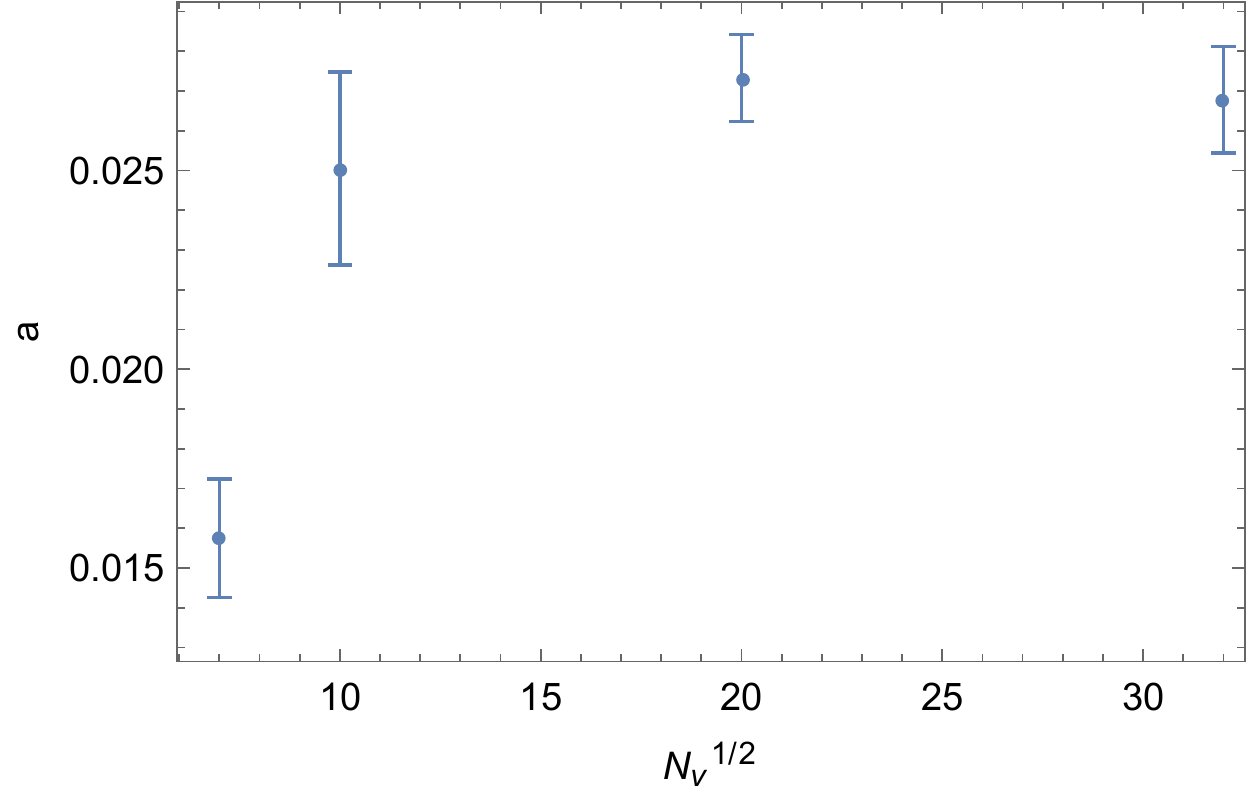}\quad
    \includegraphics[width=0.48\linewidth]{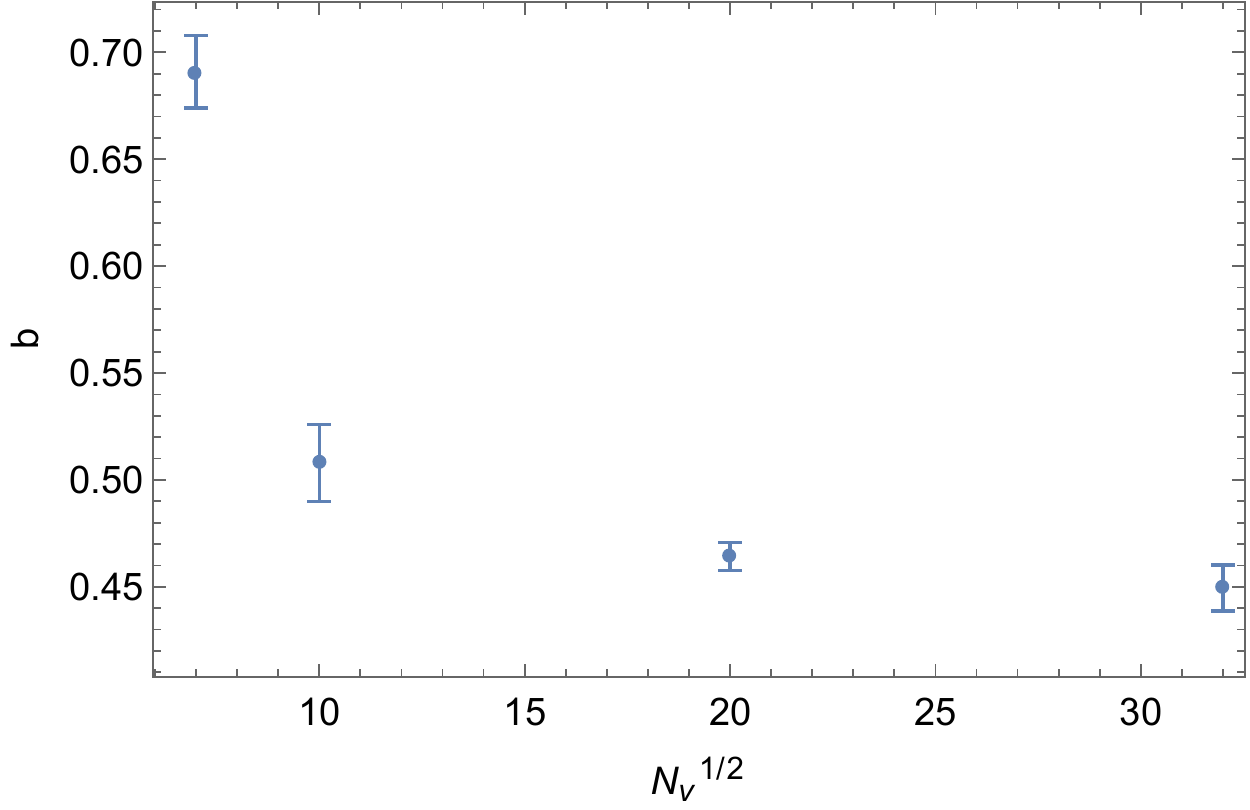}
    \caption{Dependence of fitting parameters $a,b$ on $N_v$. 
    The parameter $b$ is a monotonically decreasing function of $N_v$ (right),
    while the parameter $a$ seems an increasing function of $N_v$ (left).}
    \label{fig:coeff}
\end{figure}

Next we discuss the $N_v$ dependence.
Figures \ref{fig:min_en} and \ref{fig:coeff} show that 
the minimum energy $E_{min}$
and the fitting parameter $b$ 
are monotonically decreasing functions of $N_v$.
Then we can speculate $b\to 0$ in the $N_v\to\infty$ limit,
which means that 
a curve in the $N_{temp}$--$E_{min}$ plane (in Fig.\,\ref{fig:min_en}) becomes flat,
and we obtain $E_{min}\to -2e^{-a}$ in this limit.

Unfortunately, we cannot estimate the value of $a$ in the $N_v\to\infty$ limit,\footnote{We wish we could have fitting functions for $a$ and $b$, but we don't have sufficient number of data points in Fig.\,\ref{fig:coeff}.}
but we can discuss it in the following way.
For $N_{temp}<100$ and $N_v\geq 10^2$,
Fig.\,\ref{fig:min_en} shows that
the minimum energy in all the cases is $E_{min}\sim -1.7$,
which corresponds to $T\sim T_c$.
This is a consistent result with our previous studies~\cite{Iso:2018yqu,ShibaFunai:2018aaw}
but quite different behavior from the fitting function (\ref{eq:fit}).
This is in fact why we don't use the data points with $N_{temp}<100$
for the fitting to Eq.\,(\ref{eq:fit}).
Then we can assume that,
at least for $30\leq N_{temp}<100$,
the minimum energy $E_{min}\sim -1.7$ in the $N_v\to\infty$ limit.
Together with our speculation $b\to 0$, 
we can conjecture that $E_{min}\to -1.7$ in the limit of $N_v\to\infty$ with $N_{temp}\,(\geq 30)$ fixed.

This conjecture means that the fitting parameter $a\to 0.16$ in the $N_v\to \infty$ limit.
This might be possible since in Fig.\,\ref{fig:coeff}
the parameter $a$ seems an increasing function of $N_v$, 
but it needs to be checked in a future work.

\subsection{Conjecture for large $N_{temp}$ and $N_v$}
\label{sec:conjecture}

Summarizing the previous subsections,
we list here our conjecture:

\begin{enumerate}
\item For $N_{temp}\geq 300$, $N_h$ with the minimum energy of the RBM fixed point is given as $N_{h,min} = (0.4)^2N_v$.

\item In the limit of $N_{temp}\to\infty$ with $N_v$ fixed and $N_h=N_{h,min}$,
the RBM fixed point is at $E\sim 0$, or $T\to \infty$.

\item In the limit of $N_v\to\infty$ with $N_{temp}\,(\geq 30)$ fixed and $N_h=N_{h,min}$,
the RBM fixed point is at 
the phase transition point $T\sim T_c$, or $E\sim -1.7$.
\end{enumerate}

In other words, if we would like to find the RBM fixed point around the phase transition point, we need to generate the spin configurations with large enough size $N_v$
or with small enough number of temperatures $N_{temp}$,
and to set the number of hidden neurons $N_h\sim N_{h,min}$.
Since our previous studies~\cite{Iso:2018yqu,ShibaFunai:2018aaw} meet such criteria by chance, we found luckily such an interesting phenomenon.

\subsection{Supporting evidence for the conjecture}
\label{sec:3.4}

In our conjecture, the item 2 is easy to understand.
For larger $N_{temp}$ (with $N_v$ fixed), 
the training data include more random configurations with $E\sim 0$.
We can check it by looking at the averaged energy of training data
(in the right panel of Fig.\,\ref{fig:en_configs}).
If only random configurations are input for training an RBM, 
the loss function (KL divergence) is never reduced
and the reconstructed configurations are random ones.
Therefore, in the $N_{temp}\to\infty$ limit with $N_v$ fixed, 
the reconstructed configurations become random-like ones
and 
the RBM fixed point should be at $E\sim 0$, or $T\to\infty$.

The item 3 is more mysterious, but
Fig.\,\ref{fig:min_en} clearly shows that 
the RBM fixed point is at lower energy 
for larger size $N_v$ (with $N_{temp}$ fixed).
Let us look again at the averaged energy of training data.
Then we find that the error bars for larger size $N_v$ are apparently smaller, which means that the training data include less random configurations (with $E\sim 0)$.
The less random configurations we input, the more non-random patterns the RBM can learn.
As a result, the reconstructed configurations have more non-random patterns and lower energy.

\begin{figure}[t]
    \centering
    \includegraphics[width=0.49\linewidth]{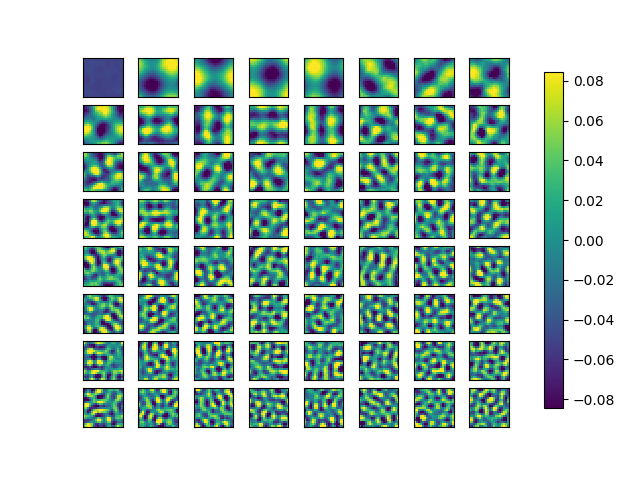} 
    \includegraphics[width=0.49\linewidth]{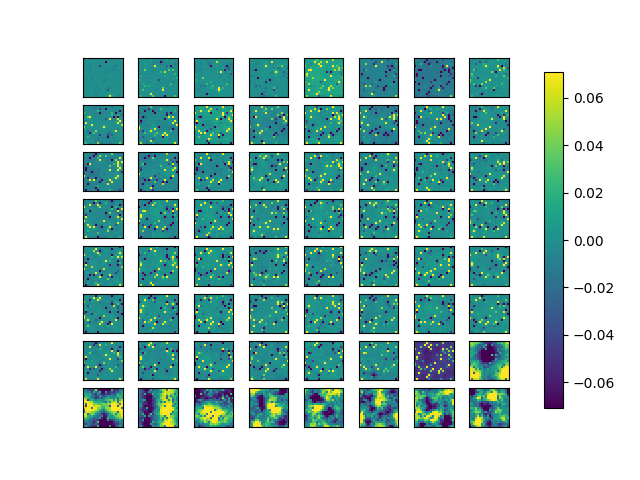}
    \caption{Eigenvectors of weight matrix $WW^T$ in the cases of $N_v=20^2$, $N_h=8^2$, $N_{temp}=100$ (left) and $700$ (right).
    They are arranged in descending order of eigenvalues from left to right and then from top to bottom.}
    \label{fig:eigen}
\end{figure}

\begin{figure}[t]
    \centering
    \includegraphics[width=0.7\linewidth]{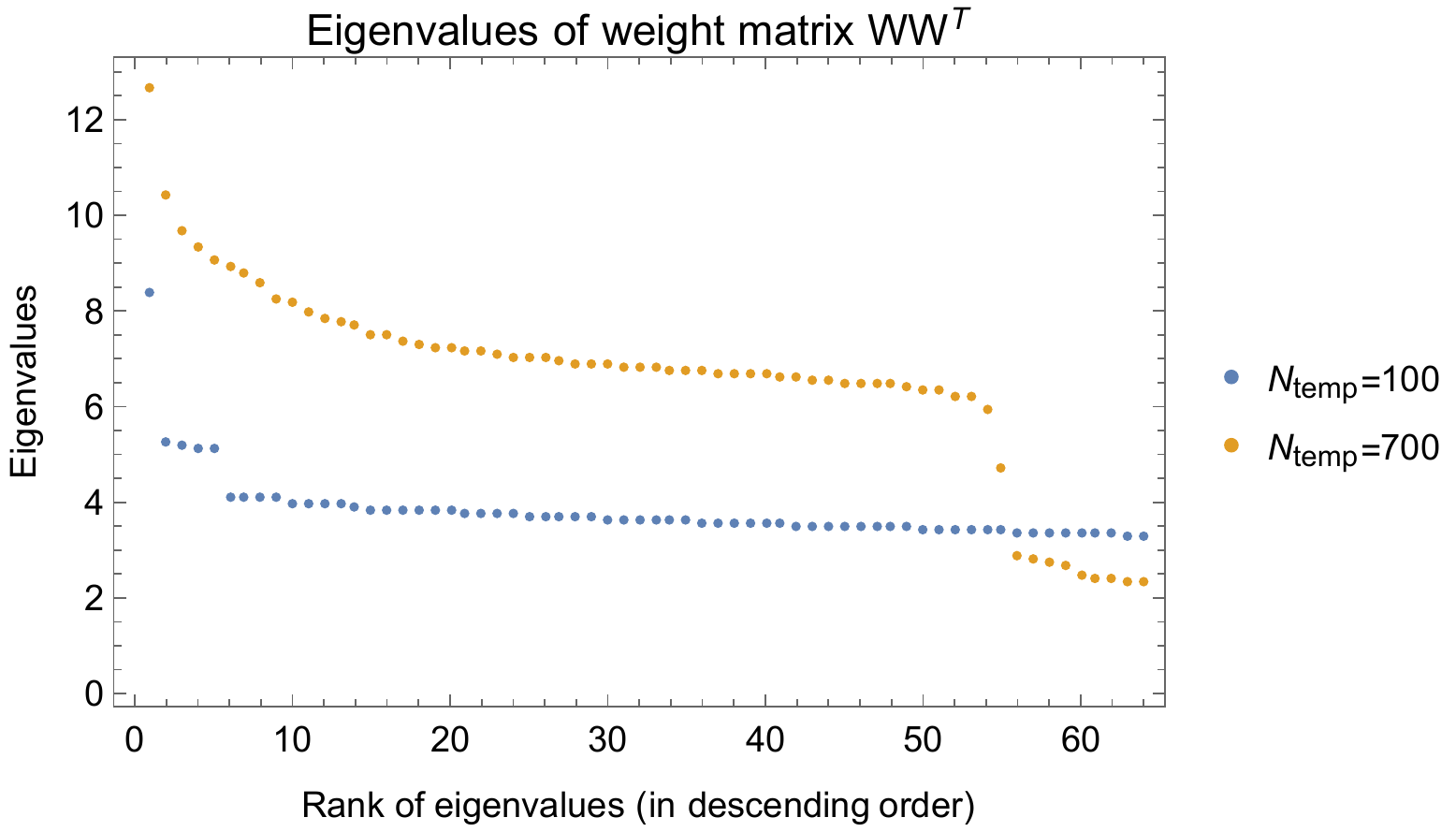} 
    \caption{Eigenvalues of weight matrix $WW^T$ in the cases of $N_v=20^2$, $N_h=8^2$, $N_{temp}=100$ and $700$.}
    \label{fig:eigenvalue}
\end{figure}

In order to check the patterns which the RBM learned,
we can 
study the eigenvectors of the weight matrix $W_{ia}$.
More precisely, we analyze the product of weight matrix
$\sum_a W_{ia}W_{aj}^T$\,,
which is independent of the basis of the hidden neurons.
Its eigenvectors in the cases of 
$N_v=20^2, N_h=8^2$, 
$N_{temp}=100$ and $700$
are shown in Fig.\,\ref{fig:eigen},
arranged in descending order of the (absolute values of) eigenvalues.
Note that there are $N_h$ eigenvectors with nonzero eigenvalues since $N_v\geq N_h$ is satisfied in all of our cases.
Then we find that 
all the eigenvectors in the $N_{temp}=100$ case have non-random patterns,
while only the last nine eigenvectors have non-random patterns in the $N_{temp}=700$ case.

The corresponding eigenvalues are shown in Fig.\,\ref{fig:eigenvalue}.
In the $N_{temp}=700$ case, 
there is a big gap between the last nine points and the others, 
which corresponds to a boundary between the eigenvectors with random patterns and non-random patterns.
On the other hand, in the $N_{temp}=100$ case,
there are no such gaps.
Except the first five eigenvectors with especially large-scale patterns,
all the other eigenvectors have patterns with various scales
and their eigenvalues are not exactly the same but close to each other.
Due to this kind of (approximate) scale invariance,
these non-random patterns appear in the reconstructed configurations
in a scale-invariant way,
and the configurations at the RBM fixed point look like those around the phase transition point $T\sim T_c$.
Naively, this should be why the RBM fixed point is around the phase transition point
in this case. 

\begin{figure}
    \centering
    \includegraphics[width=0.7\linewidth]{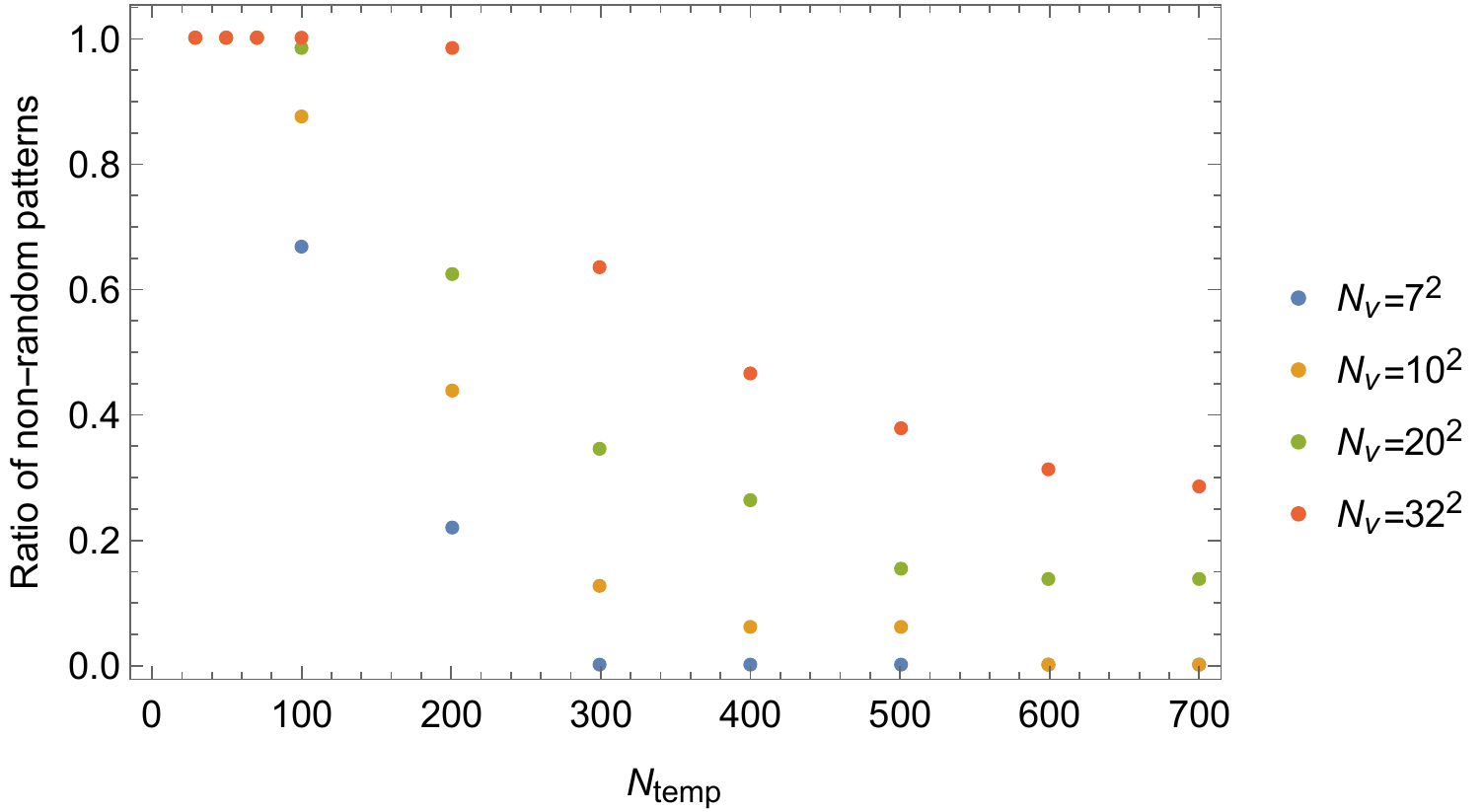}
    \caption{Ratio of the number of eigenvectors with non-random patterns in all the $N_h$ eigenvectors. 
    We fix $N_h=N_{h,min}$, {\em i.e.,} $N_h$ with the minimum energy of RBM fixed point.
    In all the cases with $N_{temp}<100$, the ratio is $1$.}
    \label{fig:num_pattern}
\end{figure}

Then, in order to discuss how the patterns which the RBM learns
depend on the parameters $N_{temp}, N_v$,
we next 
study the ratio of the number of eigenvectors with non-random patterns
in all the $N_h$ eigenvectors when we fix $N_h=N_{h,min}$,
which is shown in Fig.\,\ref{fig:num_pattern}. 
We can find that the ratio becomes smaller for larger $N_{temp}$, 
while it becomes larger for larger $N_v$.
Based on this property, we expect that 

\begin{itemize}
\item In the limit of $N_{temp}\to\infty$ with $N_v$ fixed and $N_h=N_{h,min}$,
all the eigenvectors become random-like patterns
({\em i.e.,} the ratio approaches 0).

\item In the limit of $N_v\to\infty$ with $N_{temp}$ fixed and $N_h=N_{h,min}$,
all the eigenvectors become non-random patterns
({\em i.e.,} the ratio approaches 1).
\end{itemize}
The first claim corresponds to the item 2 of our conjecture 
(in Sec.\,\ref{sec:conjecture}).
Moreover, the second claim should correspond to the item 3,
if the eigenvectors have patterns with various scales
and most of their eigenvalues are close to each other,
as discussed in the previous paragraph.
This condition 
is satisfied in all the cases of $N_v=32^2, N_{temp}\leq 100$, but needs to be checked for larger $N_v$ in a future work.

Finally, we comment on the item 1.
Through a similar discussion in Fig.\,\ref{fig:num_pattern},
we can find that 
if $N_h$ becomes smaller (with $N_v$ and $N_{temp}$ fixed), 
the ratio of the eigenvectors with non-random patterns monotonically increases.
However, this does not mean that the smallest $N_h$ equals to $N_{h,min}$.
In the region of $N_h<N_{h,min}$,
as we mentioned in Sec.\,\ref{sec:Nh}, 
the expectation values $\langle\tv_i\rangle$ is not always close to $\pm 1$
and this causes random noise 
in the reconstructed configurations $\tv_i$.
In such cases, the RBM learns only blurred patterns and 
the eigenvectors of the weight matrix $WW^T$ also have unclear patterns.
Therefore, this ``unclearness'' must be considered
besides the ratio of the eigenvectors with non-random patterns,
when we try to analytically calculate the value of $N_{h,min}$.
The author has no good idea at this time but keeps challenging this problem.

\section{Conclusion}
\label{sec:4}

We perform machine learning of the RBM to extract features of spin configurations in two-dimensional Ising model.
We find that the RBM flow of iterative reconstructions has the fixed point
in the parameter space of temperature,
which should describe nothing but the extracted features.
As shown in our previous papers~\cite{Iso:2018yqu,ShibaFunai:2018aaw},
in some cases the RBM fixed point is 
at the critical temperature $T=T_c$ in the Ising model,
although the RBM has no prior knowledge of the phase transition.

Then, in this paper, we study the dependence of the RBM fixed point on the following three parameters.
\begin{itemize}
\item $N_v$\,: the size of configurations $=$ the number of visible neurons in the RBM
\item $N_h$\,: the number of hidden neurons in the RBM 
\item $N_{temp}$\,: the number of temperatures of configurations 
\end{itemize}
Based on this dependence,
we conjecture the condition where 
the RBM fixed point is at phase transition point $T=T_c$\,:
When the number of temperatures $N_{temp}$ is fixed
and we look at only $N_h=N_{h,min}$ 
({\em i.e.,} $N_h$ with the minimum energy of the RBM fixed point),
the RBM fixed point approaches the phase transition point $T=T_c$
if the size of configurations $N_v$ becomes large enough.

We also provide the supporting evidence for the conjecture.
If the size $N_v$ becomes larger,
the number of random configurations (with $E\sim 0$) in training data
becomes smaller.
Then the RBM can learn more non-random patterns.
These patterns can be described as the eigenvectors of the weight matrix $WW^T$, and we can check that 
the number of eigenvectors with non-random patterns increases
if the size $N_v$ becomes larger.
In addition,
the eigenvectors have non-random patterns with various scales
and their eigenvalues 
are close to each other.
Due to this approximate scale invariance in the RBM weight matrix,
the configurations at the RBM fixed point should be similar to
the scale-invariant configurations at the phase transition point $T=T_c$.

Let us here comment on Ref.\,\cite{CA}.
They claim that geometrical information of the configurations is 
not in the RBM but in the NN thermometer,
since they obtain the same RBM flow even if they use a random weight matrix of the RBM with the same distribution as the trained RBM.
However, we don't use the NN thermometer in this paper and reproduce the same result in the previous studies.
Moreover, the eigenvectors of the RBM weight matrix show
the non-random and random-like patterns which the RBM learns, and also 
the eigenvalues suggest that approximately scale-invariant configurations can be reconstructed at the RBM fixed point.
Then we can claim that geometrical information of the configurations should be in the RBM.
This may suggest that the NN thermometer can measure temperature
of the configurations 
by using quantities without geometrical information,
such as the magnetization $\left|\sum_i \tv_i\right|/N_v$.

Finally, we should say that making a conjecture is not the end of the story.
In particular, the analytic calculation of $N_{h,min}$ is
an important matter to understand when and why the machine learning works well.
The author hopes to clarify it in future studies.

\subsubsection*{Acknowledgment}

The author would like to thank Jonathan Miller and Reuven Pnini for collaboration in the early stage of this work,
and also thank Satoshi Iso for his useful comments.

\end{document}